\documentclass[sigconf, screen]{acmart}

\def\eg{\emph{e.g., }} 
\def\ie{\emph{i.e., }}

\newcommand{\pquotes}[1]{\textcolor[gray]{0.35}{\textit{#1}}}

\AtBeginDocument{%
  }

\copyrightyear{2026}
\acmYear{2026}
\setcopyright{cc}
\setcctype{by}
\acmConference[HRI '26]{Proceedings of the 21st ACM/IEEE International Conference on Human-Robot Interaction}{March 16--19, 2026}{Edinburgh, Scotland, UK}
\acmBooktitle{Proceedings of the 21st ACM/IEEE International Conference on Human-Robot Interaction (HRI '26), March 16--19, 2026, Edinburgh, Scotland, UK}
\acmDOI{10.1145/3757279.3785645}
\acmISBN{979-8-4007-2128-1/2026/03}

\begin{document}

\title{Plant-Inspired Robot Design Metaphors for Ambient HRI}

\author{Victor Nikhil Antony}
\affiliation{%
  \institution{Johns Hopkins University}
  \city{Baltimore}
  \country{USA}
}
\author{Adithya R N}
\affiliation{%
  \institution{Johns Hopkins University}
  \city{Baltimore}
  \country{USA}
}

\author{Sarah Derrick}
\affiliation{%
  \institution{Johns Hopkins University}
  \city{Baltimore}
  \country{USA}
}

\author{Zhili Gong}
\affiliation{%
  \institution{Rice University}
  \city{Houston}
  \country{USA}
}

\author{Peter M. Donley}
\affiliation{%
  \institution{Johns Hopkins University}
  \city{Baltimore}
  \country{USA}
}

\author{Chien-Ming Huang}
\affiliation{%
  \institution{Johns Hopkins University}
  \city{Baltimore}
  \country{USA}
}

\renewcommand{\shortauthors}{Antony et al.}

\begin{abstract}
Plants offer a paradoxical model for interaction: they are ambient, low-demand presences that nonetheless shape atmosphere, routines, and relationships through temporal rhythms and subtle expressions. In contrast, most human–robot interaction (HRI) has been grounded in anthropomorphic and zoomorphic paradigms, producing overt, high-demand forms of engagement. Using a Research through Design (RtD) methodology, we explore plants as metaphoric inspiration for HRI; we conducted iterative cycles of ideation, prototyping, and reflection to investigate what design primitives emerge from plant metaphors and morphologies, and how these primitives can be combined into expressive robotic forms. We present a suite of speculative, open-source prototypes that help probe plant-inspired presence, temporality, form, and gestures. We deepened our learnings from design and prototyping through prototype-centered workshops that explored people’s perceptions and imaginaries of plant-inspired robots. This work contributes: (1) Set
of plant-inspired robotic artifacts; (2) Designerly insights on how people perceive plant-inspired robots; and (3) Design consideration to inform
how to use plant metaphors to  reshape HRI.
  
\end{abstract}





\maketitle

\begin{figure}[t]
\centering
\includegraphics[width=\columnwidth]{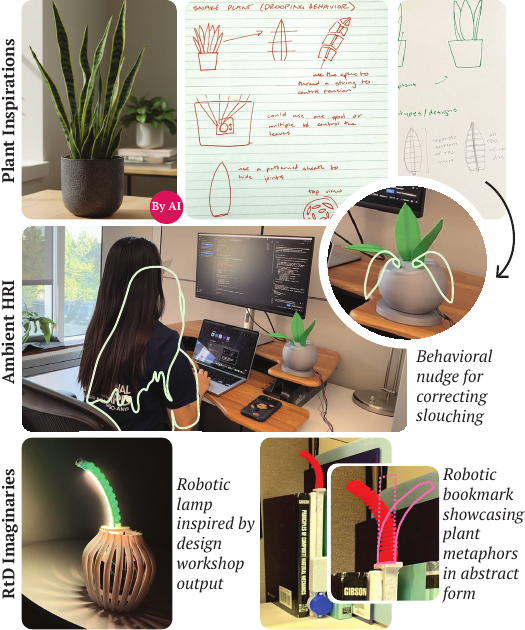}
\caption{Using a Research-through-Design process, we translate plant-inspired metaphors into robot prototypes and HRI imaginaries to illustrate the potential of plant metaphors to reshape HRI as a \textit{ambient, gradual} and \textit{interpretive}.}
\label{fig:teaser}
\end{figure}

\section{Introduction}

Plants have long been woven into human life, shaping environments, sustaining rituals \cite{ghorbanpour2017importance}, and cultivating relationships in ways that extend beyond their material forms \cite{schaal2019plants, niazi2023people}. They nourish us, adorn our spaces, and carry cultural symbolism, all while remaining largely in the background. Through their ambient presence, plants invite care and attention, foster attachment, and communicate---signaling health, neglect, season, and time through their growth and form. Plants, thus, model a paradoxical role: low-demand presences that profoundly shape atmosphere and routines through unique temporal rhythms and intuitive expressions.

Robots are increasingly being woven into everyday life, taking on roles ranging from care \cite{alves2015social,lee2018reframing, antony2025social} to collaboration \cite{sullivan2024making}. Yet, their design and interaction models have overwhelmingly drawn on anthropomorphic \cite{kalegina2018characterizing, kahn2008design} and zoomorphic sources \cite{sauer2021zoomorphic, antony2025social}. These inspirations often produce overt, high-demand forms of interaction, regardless of the role. Plants offer a simultaneously familiar and underexplored model for HRI. They exemplify forms of presence, temporality, and relationality that diverge from dominant paradigms in robotics, providing a rich set of metaphors for reimagining HRI as ambient and gradual yet intuitively expressive.

To approach this speculative design space, we adopted a Research through Design (RtD) methodology \cite{zimmerman2007research, zimmerman2010analysis} to explore how plant metaphors might inform new directions in HRI. RtD enabled us to probe ``what is the right thing to design'' using plant metaphors \cite{alves2021collection}. Through iterative cycles of brainstorming, prototyping, and building, we examined \textit{what design primitives could emerge from plant metaphors and morphologies} (RQ1), and \textit{how these primitives might be combined to construct expressive robotic forms} (RQ2). The resulting prototypes were not intended as fully functional systems but as speculative artifacts that open new imaginaries for how robots might mediate relationships. To extend this inquiry, we conducted design workshops to elicit participants’ imaginaries and examine \textit{how do people perceive plant-inspired robots} (RQ3) \cite{ostrowski2020design}. We consolidated our workshop findings into annotated portfolios \cite{hoggenmuller2021eliciting} and our learnings from the full RtD process into design considerations to guide future of plant-inspired HRI.

This work makes the following contributions:
\begin{enumerate}
\item \textbf{Artifacts}: A suite of open-source\footnote{link to materials: https://tinyurl.com/3k82nww3} robotic plant prototypes developed through our research through design practice exploring expressive potential and ambient interaction.
\item \textbf{Designerly Insights}: A qualitative understanding of how people perceive and imagine plant-inspired robots through design critiques, annotated portfolios and imaginaries.
\item \textbf{Design Considerations}: A set of design considerations to inform translation of plant-metaphors for robot design.
\end{enumerate}


\section{Related Works}

\subsection{Sources of Inspiration for HRI}
Robot designers have drawn on diverse sources of inspiration to create expressive and relatable machines. Human–human interaction remains foundational \cite{sauppe2014design}, with nonverbal cues such as gaze \cite{huang_gaze_2015}, posture \cite{nertinger2024designing}, and timing \cite{cao2025interruption} adapted into robotic behaviors to signal attention \cite{bruce2002role}, emotion \cite{antony2025xpress}, and intent \cite{gao2022evaluation}. Animation practices \cite{koike2024sprout, hu2025elegnt} and comic tropes \cite{koike2023exploring, young2007robot} have provided further inspiration, offering visual metaphors and exaggerations that allow robots to externalize internal states through motion lines, symbolic icons, or stylized gestures. Zoomorphic models have likewise shaped locomotion \cite{mo2020jumping}, control \cite{rahmani2016robust}, and embodiment \cite{pfeifer2007self}, from robotic skins mimicking goosebumps to tails signaling mood \cite{hu2023can, singh2012animal}.

Plants, as another source of bio-inspiration, have informed scattered explorations: growth-inspired locomotion \cite{del2024growing, yan2019design}, smart materials \cite{speck2023plants}, seed-like sensors \cite{jun2017plant}, plants as actuators \cite{hu2024designing} and interactive displays \cite{holstius2004infotropism, bhat2021plant, grandos2025roboplant}. These works demonstrate the technical and aesthetic promise of plant metaphors, but often focus on isolated functions or material embodiments. Similarly, somaesthetic and ambient interaction models foreground felt experience and slow feedback, but are often realized through material \cite{hook2016somaesthetic} or kinetic \cite{thompson2021ambidots} artifacts rather than robotic systems. What remains missing is a broad inquiry into how plant-inspired metaphors of ambience, form, temporality, and expressivity can reframe HRI.



\subsection{Research Through Design in HRI}
Research through Design (RtD) is a methodology in which designing, prototyping, and making are central to knowledge creation. Rather than validating what is, RtD uses artifacts as epistemic tools to probe what could be—challenging assumptions, framing new questions, and expanding design imaginaries \cite{zimmerman2007research, zimmerman2010analysis}. Through iterative cycles of creation, critique, and reflection, RtD contributes to the generation of intermediate-level knowledge that is abstract enough to guide future designs, yet grounded in practice \cite{lupetti2021designerly}. This knowledge is often communicated through designerly formats such as annotated portfolios \cite{hoggenmuller2021eliciting, lowgren2013annotated}, critical reflection \cite{ljungblad2024critical, sengers2005reflective}, and the cultivation of design imaginaries \cite{winkle2025robots, zaga2024designing}.

Within HRI, RtD has emerged as a way to question dominant assumptions and expand the landscape of robotic possibilities. It has helped reframe what kinds of robots are worth building, moving beyond purely functional evaluation toward aesthetic, affective, and cultural modes of understanding \cite{luria2019championing, luria2021research}. Design-led inquiries have surfaced new interaction paradigms, from intimate, socially assistive robots \cite{isbister2022design, hoggenmuller2021eliciting}, to lucid urban robots \cite{hoggenmuller2021eliciting} to speculative engagements with alternative AI futures \cite{Lindley2020AIDesign, Benjamin2024GenAIRtD, hu2025elegnt}. RtD is well suited to explore how the breadth of plant metaphors may reframe HRI as a whole, rather than only shaping isolated design elements. By making and reflecting with speculative plant-inspired artifacts, we surface new possibilities, and consolidate these learnings in designerly formats to seed HRI futures enriched with plant metaphors.

\section{Design Process}
In our RtD process, we treated plants as metaphoric resources rather than blueprints, using their qualities of presence, temporality, morphology, and behavior to frame inquiry. We, first, developed a set of plant-inspired design primitives (\ie \textit{leaves, stems, flowers}, and \textit{pots}) that scaffolded exploration of behavioral and morphological potential (\textit{RQ1}). Then, we extended these primitives into speculative prototypes to act as probes and imaginaries of how robots might inhabit everyday life with the subtlety, ambience, and relationality of plants (\textit{RQ2}). Lastly, We examined how people interpreted, and reimagined plant-inspired robots through design workshops (\textit{RQ3}).


\begin{figure*}[t]
\centering
\includegraphics[width=\textwidth]{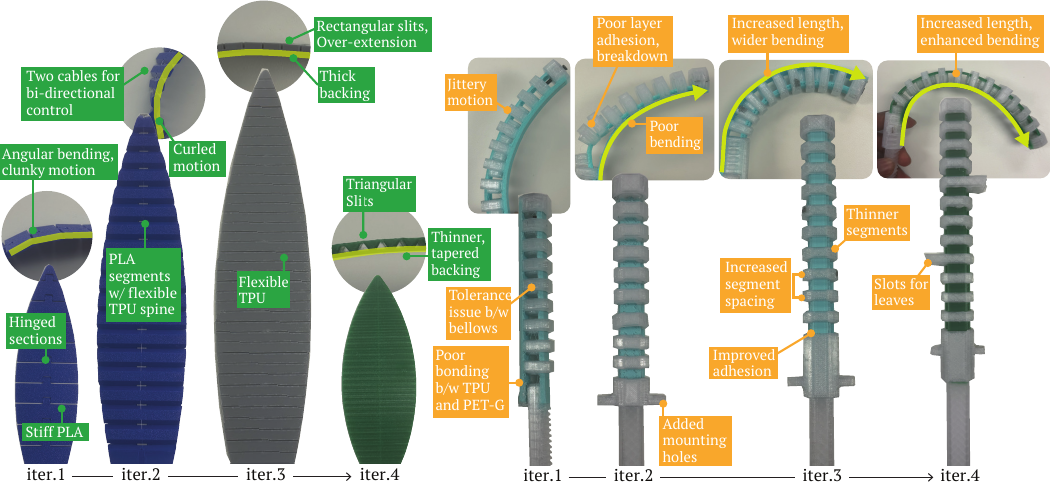}
\caption{Our \textit{leaf} and \textit{stem} primitives underwent several design iterations to realize plant-like expressive potential.}
\Description{This figure shows 4 iterations of the leaf primitive on the left side and 4 iterations of the stem primitive on the right side. The first leaf is small, blue with hinged section and manufactured with fully for PLA - this had angular bending and chunky motion; the second leaf was blue, but twice as large and printed with PLA but with a flexible TPU spine; It has two cables for bi-directional control which lead to curled motion. The third leaf is grey, fully printed with TPU but its rectangular slits and thick back layer lead to overextension. The last and final leaf is green and features triangular slits and has thinner, taper backing. The first stem module had tolerance issues between the inner and outer bellows and poor bonding b/w TPU and PET-G leading to jittery motion; the second stem still suffered from poor layer adhesion leading to breakdown but added mounting holes; the third stem improved adhesion, increased segment length and made segments thinner, and overall length leading to wider bending; the last stem increased length more leading to enhanced bending and added slots for leaves.}
\label{fig:primitives_iterations}
\end{figure*}

\subsection{Robot Design Inspiration from Plants}

We approach plants not as blueprints to replicate but as metaphoric guides for reorienting robot design, focusing on four features: \textit{ambient presence, temporality, morphology,} and \textit{behavior}. Each feature highlights a quality of plant life that, when translated into design decisions, challenges assumptions of immediacy, overtness, and anthropomorphism in HRI.

Plants exemplify \emph{ambient presence}. They are woven into homes, workplaces, and public settings without demanding focused attention, yet they quietly shape the atmosphere and behavior. Plant ambience serves as a metaphor for robotic artifacts that exist in the background while still exerting perceptible influence—systems that support, scaffold and enrich, rather than dominate, interactions.

Plants operate in unique, distinctive \emph{temporal timelines}. Their rhythms of growth, decay, and seasonality unfold across hours, days, and months rather than simply individual moments. This temporality offer a design metaphor that encourages gradual, cyclical, and accumulative interaction through less transactional and more ritualized, subtler cues that gain meaning over time.

Plants inspire a rich \emph{morphological} range. The rigidity of stems, translucency of petals, and curling of tendrils point to material and form motifs—softness, vibrancy, asymmetry, stretchability—that can evoke plant-like sensibilities even without literal mimicry.

Plants offer a rich \emph{behavioral} metaphors for ambient cues. Gestures such as opening and closing, swaying, drooping, or slow expansion communicate state and invite interpretation despite their subtlety. Transposed onto robot design, these gestures—along with changes in light, color, or material state—become a rich expressive repertoire for signaling, affect, and engagement.

These metaphoric lenses helped ground the design exploration that follows: an iterative process of prototyping to operationalize plant-like presence, time, form, and gestures in robotic artifacts.

\subsection{Building Plant-Inspired Robot Primitives}

Drawing on a survey of plant life, we identified \textit{leaves}, \textit{stems}, \textit{flowers}, and \textit{pots} as key elements. Treating these elements as design primitives structured our prototyping process and provided scaffolding for translating plant metaphors into tangible robotic primitives. Figure \ref{fig:primitives_iterations} illustrates the major iterations of \textit{leaf} and \textit{stem} primitives.

\textbf{Leaf.} The \textit{leaf} primitive was iteratively developed to explore leaf-inspired expressivity (see Fig. \ref{fig:primitives_iterations}). Our \textit{leaf} is based on a lightweight, string-driven mechanism. Each \textit{leaf} is additively manufactured using TPU to exploit its flexibility and stretchability. The structure consists of a thin base layer reinforced by a central spine, with triangular wedge segments between the top and base layers; these facets bias the leaf into a naturally curled state. A narrow channel within the spine guides an actuation string from the base to an anchoring point at the tip. Pulling the string increases tension along the spine, flattening the wedge segments and straightening the \textit{leaf}; releasing the tension allows the elastic base to recover, returning the \textit{leaf} to its curled form. This mechanism design supports diverse shapes and scales while preserving expressive versatility, enabling gestures such as wilting, perking, or unfurling.

\textbf{Stem.} The \textit{stem} module was iteratively designed to explore more rigid structures and a distinct set of behaviors (see Fig. \ref{fig:primitives_iterations} for major iterations). Its design adapts a concentric bellow mechanism, originally developed for soft and continuum robots, into a compact actuation unit. The \textit{stem} achieves bending through the lateral displacement of an inner bellow encased within an outer bellow: pushing the bellow creates curvature in one direction, while pulling it bends its the other way. The \textit{stem} is additively manufactured: each bellow combines a TPU backbone with PET-G arches, with the outer bellow giving structure while allowing deformation and housing a cavity for the inner bellow to slide within. The two bellows are fastened together at the top with a small screw, allowing the inner bellow to remain constrained while still transmitting bending forces. At the base, the inner bellow integrates a rack interface, enabling precise rack-and-pinion actuation. The mechanism enables smooth, reversible bending that scales across stem-like geometries, supporting expressive gestures—leaning, bowing, swaying—that evoke the characteristic motion of plant stems.


\begin{figure*}[h]
\centering
\includegraphics[width=\textwidth]{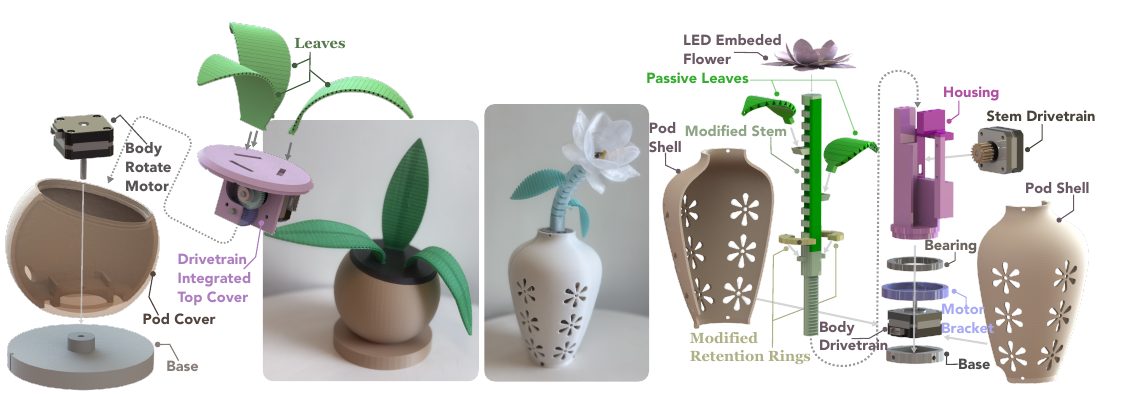}
\caption{Our prototypes leverage our primitives in distinct configurations to be probes of the plant-inspired design space.}
\Description{The there are four images in this figure. The first shows an exploded CAD of the snake plant showcasing how the base fits into the pot into the body rotate motor then the top cap with the integrated drive train featuring gears is attached and the leaves are added at the end. The second image is that of the Snake Plant prototype with green leaves, black top cap, and beige pot and base; the third image shows the Flower prototype with a white pot accentuated with vents and a teal blue stem and two leaves and translucent flower. The last images shows the exploded CAD view of the flower plant with the stem drive train fitting into the housing and the base with the rotate motor.}
\label{fig:prototypes-CAD}
\end{figure*}


\textbf{Pots and Flowers.} We designed pots not only as containers of other primitives but as extensions of expression—sites where form, color, and motion could emerge meaning together. The pots varied in size, shape, color, and materials not only for aesthetics but also for functional performance (e.g., ventilation holes for heat management, PET-G for melt resistance). Some pots turned or swayed, others moved across the surface, offering different kinds of presence. Our experiments with flowers were limited—exploring color, translucency, and the glow of embedded LEDs.

\textbf{Actuation and Materials.} Through our design practice, We explored an array of actuation methods to realize the primitives’ expressive range, focusing on controllability (\eg fidelity, frequency, speed, acceleration). Servo motors offered simplicity but proved too noisy and jittery undermining ambience and expressivity. Switching to stepper motors with TMC drivers enabled near-silent operation and fine resolution through microstepping. Iterating on motor models and adding gears and heat-management features provided sufficient torque while preserving quiet, ambient presence, allowing gestures to unfold on plant-like timescales rather than abrupt robotic motions. Across primitives, we explored material choices such as Polylactic Acid (PLA), Thermoplastic Polyurethane (TPU), and Polyethylene Terephthalate-Glycol (PET-G); their differing properties (\eg rigidity, elasticity, translucency) proved central in shaping both morphology and expressive potential.

\subsection{Prototyping Plant-Inspired Robots}

To probe this design space further, we constructed four plant-inspired robotic prototypes: \textit{Snake Plant, Flower Plant, Keeper Plant,} and \textit{Dancing Pot}. These speculative probes intended to prompt reflection on various aspects of plant metaphors in robot design. Each prototype combines the primitives in distinct configurations, evoking new imaginaries around contextual and expressive possibilities.

\textbf{Snake Plant.} Inspired by the vertical geometry of a snake plant, this prototype combines three actuated leaves with a rotatable pot. The leaves are mounted on a top cap angled relative to the base; when the pot rotates, this geometry produces a characteristic leaning motion, as if the plant is subtly shifting direction. The leaves in this probe move synchronously, enabling coordinated gestures that blend swaying, leaning, drooping and fluttering to create ambient cues. Each leaf is tensioned with slight variance to manifest natural asymmetries; our design practice revealed how independently actuated leaves, moving asynchronously, can enrich plant-like variance (see Fig. \ref{fig:prototypes-CAD} for mechanical design).

\textit{Imaginary}. Placed on a desk, the plant could mirror and respond to a user’s posture—remaining upright with good posture, drooping as posture deteriorates, and gently recovering when corrected. If posture continues to worsen, the plant could nudge the user to adjust. It would greet the user at the start of the day and perform a small celebration gesture at the end, marking daily rhythms (see Fig. \ref{fig:ap-snake}). Link to videos of all imaginaries can be found in Appendix.

\begin{figure*}[h]
\centering
\includegraphics[width=\textwidth]{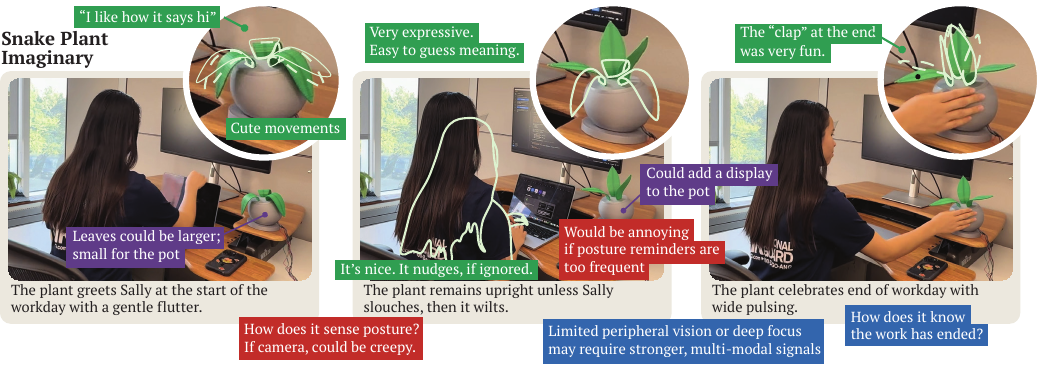}
\caption{Annotated portfolio showcasing the \textit{Snake Plant} imaginary and our participant's perceptions. Annotations are color coded: green for positive impressions, red for critiques, blue for questions or reflections, purple for concrete extension ideas.}
\Description{This figure hows an annotated portfolio with three frames from a story board; the first frame shows a person opening a laptop and a snake plant robot sitting on their desk - a zoomed circle shows the greeting pulsing with motion overlay. There are green annotations that read I like how it says high and cute movement; purple annotation that says leaves could be larger; small for the pot; red annotation that says how does it sense posture, if camera, could be creepy; The second frame of the storyboard shows a person their slouched version overlayed with green outline. the zoomed in view of the snake plant shows its drooped with same green outline. There are green annotations saying Very expressive; easy to guess meaning; its nice, it nudges, if ignored; red annotations saying would be annoying if posture reminders are too frequent. and a purple annotation saying could add a display to the pot and a blue annotation saying Limited peripheral vision or deep focus
may require stronger, multi-modal signals. The last frame depict the end of the work session with the person touching the snake plant. the zoomed in view shows the snake plant pulsing exagerated doing a clapping motion; a green annotation says the clap at the end was very fun; a blue annotation says how does it know the work has ended}
\label{fig:ap-snake}
\end{figure*}

\textbf{Flower Plant.} Inspired by the expressive aesthetics of flowering plants, this prototype centers on an actuated stem  crowned with a flower. The pot enables rotational motion, its patterned surface accentuating each turn. Along the stem, passive leaves extend the botanical metaphor, adding layers of texture and offering passive expressivity that can flutter subtly as the stem moves. In concert, stem bending and pot rotation produce composite gestures that echo the attentive shifts of a sunflower orienting itself in its environment.

\begin{figure*}[h]
\centering
\includegraphics[width=\textwidth]{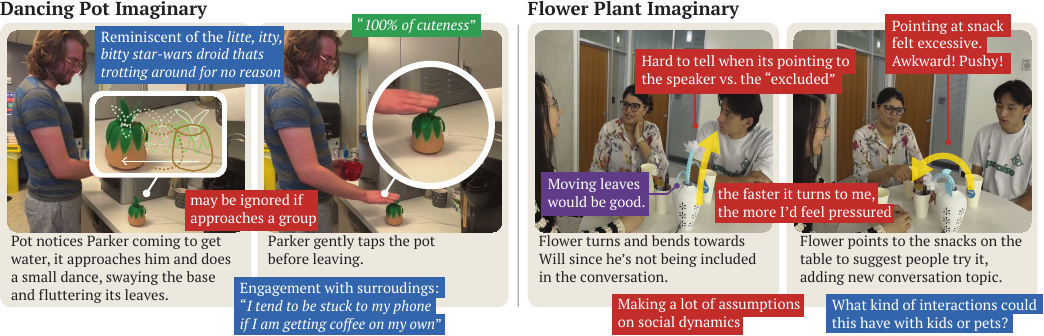}
\caption{Annotated portfolio showcasing the \textit{Flower Plant} and \textit{Dancing Pot} imaginaries and our participant's perceptions.}
\Description{There are two storyboards; the first storyboard on the left showcases the dancing pot imaginary; in the first frame: Pot notices Parker coming to get
water, it approaches him and does a small dance, swaying the base and fluttering its leaves; there is zoomed in view showing the motion through overlayed sketches. the second frame shows Parker gently tapping the pot
before leaving. There is a zoomed version showing the tap. There is a green annotation showing 100\% of cuteness and a blue annotation saying: Engagement with surroudings: I tend to be stuck to my phone if I am getting coffee on my own. The second storyboard showcases Flower Plant imaginary. The first frame depicts three people engaging in a conversation and one person not being included. Flower turns and bends towards
Will since he’s not being included in the conversation; There are red annotation saying Hard to tell when its pointing to
the speaker vs. the “excluded” and Making a lot of assumptions on social dynamics. The second frame shows the Flower points to the snacks on the table to suggest people try it, adding new conversation topic. There is a red annotation saying Pointing at snack felt excessive. Awkward! Pushy! and a blue annotation stating What kind of interactions could this have with kids or pets?.}
\label{fig:ap-flower-dancer}
\end{figure*}

\textit{Imaginary.} At a coffee table, the robot joins the conversational flow—turning toward active speakers, leaning gently toward quieter participants to invite inclusion, or gesturing toward shared objects, like snacks, to ease conversational lulls (see Fig. \ref{fig:ap-flower-dancer}).


\textbf{Keeper Plant.} Drawing in part from the Venus flytrap, this prototype explores a more function-focused design through eight small leaves—four actuated and four passive—arranged radially around a central cavity. The actuated leaves can curl inward to enclose or unfurl outward, creating a dynamic perimeter around the cavity, while passive leaves provide additional texture and asymmetry. A light sensor detects the presence of objects, informing behaviors. This configuration highlights how plant-inspired gestures of folding and blooming can take on functional roles of holding and presenting.

\textit{Imaginary.} Used as a pill organizer, the plant gently grips the box with its actuated leaves when medication is not due. At the appointed time, the leaves slowly unfurl and softly pulse as a nudge to take the medicine. Once the pill box is returned, the leaves close again, re-establishing their protective hold (see Appendix B).


\textbf{Dancing Pot.} This probe merges plant-inspired expressivity with locomotion through a mobile pot. A fern-inspired, two-tier leaf arrangement crowns the top cap: an outer ring of passive leaves forms a textured frame, while an inner ring of variably tensioned, actuated leaves produces subtle asymmetric motion. Juxtaposing the pot’s locomotion with delicate leaf gestures explores how mobile robots and plant metaphors can mutually enrich each other.


\textit{Imaginary.} Positioned near a water dispenser, the plant may approach visitors and perform a short dance as they fill their glass creating a playful sense of ambient presence (see Fig. \ref{fig:ap-flower-dancer}). 


\section{Eliciting Imaginaries Workshop}
\begin{figure*}[h]
\centering
\includegraphics[width=\textwidth]{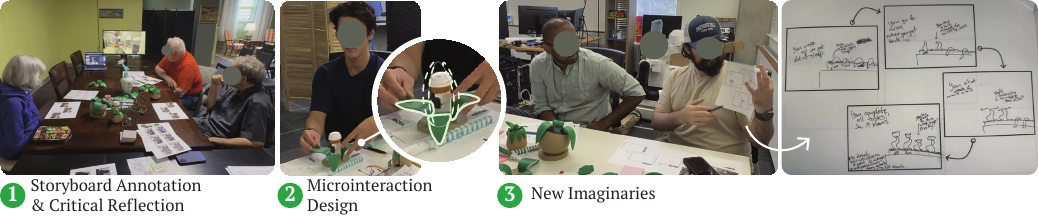}
\caption{Our design workshops elicited perceptions and new imaginaries of plant-inspired robots with three phases of activities. }
\Description{There are five panels depicting the three phases of the design workshop; THe first image depicts phase 1: storyboard annotation and critical reflection and shows  three older adults sitting around brown, wood table and looking a video playing on a screen; The next two panels depict phase 2 which is Microinteraction design and shows a young man manipulating the top cap of the Snake Plant to completely enclose the pillbox and a zoomed in view shows the movement via overlay. The last two images depict the last phase: New Imaginaries -- there are two people in the image with one person looking at the other as he points as a paper with a storyboard and there is a zoomed in version of a storyboard in the final image.}
\label{fig:process}
\end{figure*}

To explore how plant-inspired robotic forms might be interpreted and re-imagined, we conducted three 90-minute design workshops structured in three phases as approved by our Institutional Review Board. The workshop combined speculative scenarios, hands-on engagement with prototypes, and collaborative design activities to elicit participants' critical reflections (see Fig. \ref{fig:process}) \cite{antony2023co, mott2022robot}. 

\textbf{Participants}. To gain wide perspectives and design insights, we recruited 11 participants with diverse backgrounds [(6M, 5F), ages 19–82 ($M=39.7, SD=24.8$)]; compensation was \$25. Workshop \#3 consisted entirely of older adults. Prior experience with robots ranged from  research backgrounds (n=4) to coursework exposure (n=1) and limited encounters (\eg exhibitions) (n=6). Plant experience was mostly casual (n=8), with some regular gardening (n=3). Creative expertise varied: professional or academic (n=3; \eg product design, urban planning), regular creative practice (n=4; \eg painting, music), hobbies (n=3). Full demographics are provided in the appendix. Quotes are reported with participant IDs (\eg, P1).



\textbf{Workshop Process.} 
After providing informed consent, participants were engaged in the three phases: 1) Storyboard Annotation and Critical Reflection, 2) Micro-Interaction Design, and 3) New Imaginaries.
They were encouraged to physically interact with the prototypes, touching, moving, and manipulating them to connect the observed behaviors to material qualities. Our prototypes and primitives were available as shared resources.

\textit{Storyboard Annotation \& Critical Reflection.} We began by presenting four video storyboards each depicting an imaginary for a prototype. Each video was paired with a printed storyboard. Participants were asked to annotate the storyboards with their reactions (\eg impressions, ideas, questions). After every two videos, we facilitated a group discussion in which participants shared annotations and reflected on what felt expressive, what seemed confusing, and what they would change. Afterwards, participants engaged in a critical reflection guided by the following prompts: what expressive or interaction modalities would resonate most; in what contexts or applications would such robots be valuable; and when would plant-inspired robots feel intrusive rather than supportive.

\textit{Micro-Interaction Design.} Participants worked in pairs to either modifying one of the presented storyboards, or to sketch a new interaction sequence altogether. A three-panel storyboard: trigger, robot action, user reaction was provided to support design process. Pairs were encouraged to puppet the prototypes or ``\textit{bodystorm}'' \cite{segura2016bodystorming, antony2025onbodydesign} the sequence to illustrate expressive timing and gesture. This activity was skipped in Workshop \#3 due to time constraints.

\textit{New Imaginaries.} Building on their discussions, participants imagined new contexts and selectively sketched speculative scenarios for plant-inspired robots, envisioning morphology, temporality, behavior, and expressions. These sketches moved beyond immediate impressions and individual interactions, opening up new imaginaries of how plant metaphors could reorient HRI. 

\textbf{Data Analysis.} 
All workshop audio was transcribed and analyzed alongside design materials (\eg annotated storyboards). Relevant video segments of bodystorming were reviewed to examine gestures, timing, and puppeteered expressions. We conducted thematic analysis \cite{kiger2020thematic}, with two authors coding the data and resolving disagreements by consensus; We identified themes such as intrusiveness, expressivity, ambiguity, and aesthetics and we organized plant-inspired HRI imaginaries that emerged in the workshops.
Insights from annotations, discussions, and speculative sketches, combined with our RtD learnings, converged into a set of design considerations for translating plant metaphors to robot design.


\textbf{Post-Workshop.} Participants imagined non-literal robot designs, which led us to prototype a plant-inspired lamp and bookmark (see Fig. \ref{fig:teaser}) to enrich our RtD insights. To consolidate perspectives, we collated annotated storyboards into portfolios for each prototype (see Fig. \ref{fig:ap-snake}, \ref{fig:ap-flower-dancer}, and Appendix B), capturing how participants perceived, critiqued, and reimagined plant-inspired robots. 




\section{Imagining Plant-Inspired Robots}

We extracted participant imaginaries from phases 2 and 3 of the design workshops to understand the breadth of contexts envisioned for plant-inspired robots. These imaginaries reconfigured and built upon our prototypes, framing plant-inspired robots as \textit{ambient, expressive,} and \textit{multifunctional} artifacts woven into everyday routines. Below, we share illustrative examples of imaginaries that emerged:


\textbf{Mediating Social Interactions.} 
While participants were generally critical of robots managing conversational dynamics as in the \textit{Flower Plant} imaginary, they reconfigured the concept toward more structured settings (\eg classrooms) and subtler modes of mediation (\eg reflecting shared states). For instance, participants imagined a set of classroom plant robots wilting as noise levels rose, inviting quieter behavior through a metaphor of communal care.

Extended to smaller interpersonal contexts, particularly parent-child communication, participants imagined plant-like robots using posture (\eg drooping, curling, perking up) to signal an individual’s mood. Each family member might have their own robot, forming a shared garden that quietly externalizes the emotional atmosphere of the household. These states could be either sensed automatically or set intentionally, offering users—especially those who struggle with verbal expression—a non-verbal channel for emotion and helping to reduce friction caused by unspoken feelings. 

Imagining this concept in long-distance relationships, participants speculated on ``twin'' plant robots, paired but in separate locations, could mirror one another’s behavior. A gentle sway or rhythmic pulse could indicate presence allowing people to share simple moments across space. See \S\ref{sec:design_guidelines} for expression design insights.

\textbf{Work and Focus.} Participants extended the \textit{Snake Plant} slouch-correction imaginary to broader work and focus contexts. Rather than limiting interaction to a single corrective behavior, they envisioned the robot could signal a wider range of wellbeing cues, such as reminding users to move after sitting too long, to rest their eyes, or to refocus when distracted using plant-like behaviors.

Participants also imagined these robots as expressive tools for remote collaboration. One speculative concept featured a flower-like robot with a camera nestled inside its bud, which could open and orient toward a user’s face, point of gaze, or gestures adding embodied nuance to video calls. Outside of active use, its flower petals could open slightly to indicate availability, closing to mark ``do not disturb''; Its upright posture would enable physical privacy by enclosing and pointing the camera away when not in use \cite{tang2022confidant, rueben2016privacy}. 


\textbf{Sleep and Time Hygiene.} Participants envisioned plant-inspired robots embedded into everyday domestic routines, particularly around sleep hygiene and transitions across the day. P11 imagined a plant-inspired lamp robot functioning as a reading lamp (see Fig. \ref{fig:teaser}), offering utility while subtly shaping nightly rituals. As evening progressed, the robot  would gradually dim and shift its posture, signaling the close of the day and gently nudging the user toward rest. Rather than delivering explicit reminders, the robot’s slow, ambient behavior would serve as a quiet prompt for winding down. Participants suggested that plant robots could express ambient conditions (\eg time of day, weather, or air quality) through changes in posture, or color, supporting subtle environmental awareness.

\textbf{Habit Tracking.} A set of plant robots, each tied to a specific habit or behavior, was proposed by P5 to support habit and task tracking. For example, one plant might grow when a run is completed, while another droops when journaling is skipped. Together, the collection formed a robotic garden whose overall vitality and liveliness reflected the user’s self-care across domains. Instead of numerical logs or app notifications, progress was externalized through expressive, plant-like growth and decay, encouraging interpretation and emotional investment in the ongoing cultivation of habits.

\section{Translating Plant Metaphors for Ambient HRI }\label{sec:design_guidelines}

We consolidated reflections from the RtD process and insights from the design workshops into a set of design considerations for translating plant metaphors into robotic artifacts. Organized along four axes—\textit{Presence, Palette, Rhythms,} and \textit{Shape}—these considerations illuminate how different aspects of plant inspiration can guide the design of robots and reframe HRI more broadly.

\subsection{Situating Presence}
The role, placement and ecology of plants within a space shape how people live with them and offer cues for situating the presence of robots as everyday artifacts.

\textbf{Multiplicity of Role.} Our speculative prototypes, presented in narrow imaginaries, surfaced critical questions of value and necessity. As P9 remarked, \pquotes{``I don’t want more clutter. I have enough on my desk...\textbf{why} do I need another thing?''}. Plants offer a useful lens for this critique: their presence—at once decorative, ritualistic, ecological—with a limited footprint is easily justified. Likewise, robots could embody many roles simultaneously: a ritual partner, a decorative presence, and a component of a lively environment. Translating the multiplicity of plant functions into compelling robotic artifacts that contribute tangible value to lived environments proportional to their presence can help address such criticisms.



\textbf{Cultivation}. Plants are not static; people prune, graft, arrange, and repot them to suit their needs and environments. Participants similarly imagined adapting robots to their own routines—choosing which behaviors to track, and defining how the robot should respond to particular events. Supporting such personalization points toward the need for easy-to-use end-user tooling. Generative AI could help non-experts sketch interactions \cite{antony2025id}, prototype behaviors \cite{antony2025xpress}, and refine execution \cite{karli2024alchemist} while grounding choices in the robot's characteristic presence, temporality, and expressiveness. In this way, robots could, like their botanical counterparts, be cultivated over time into functional yet personally meaningful forms.

\textbf{Ecology}. Plants rarely exist in isolation; they gather into gardens, groves, and forests whose collective forms and rhythms create layered atmospheres. Likewise, robots could operate as ecologies—distributed networks whose behaviors coordinate or contrast to produce emergent meaning \cite{glas2012network}. Such robotic collectives emerged in our design workshops to mirror group states and distribute roles across artifacts; embodying forms of shared intelligence that no single robot could achieve. In our design practice, we explored low-fidelity sensing (\eg light, touch) to reinforce a sense of ambient presence and avoided higher-resolution sensing (\eg cameras) that risk intrusiveness. Yet many imagined contexts—from collaborative work to household wellbeing—demanded richer contextual awareness. This raises open questions around how to cultivate plant-like distributed intelligence: fusing diverse signals (environmental monitors, health data) collected using less-intrusive sensors \cite{zhao2018through} and processing them locally \cite{tang2022confidant}. Leveraging plant-like ecologies for robotic collectives can enable new form of expressivity, utilities and privacy-preserving sensing.


\subsection{A Rich Yet Delicate Expressive Palette}

Plants offer a distinctive repertoire of gestures (\eg swaying, drooping) that shift robotic expression away from the directness of animal or human models toward more ambient, interpretive forms. Yet, using this expressive palette requires striking a delicate balance.


\textbf{Calibrating Meaning.} Plant gestures are rarely fixed in meaning. Where a wagging tail reads unambiguously as arousal \cite{siniscalchi2018communication}, a drooping leaf might signal fatigue, sadness, or simply growth in a new direction. This openness of interpretation can be poetic, inviting users to read their own meanings into behaviors, but also risks confusion. In our workshops, the \textit{Keeper Plant}’s pulsing was seen as a gentle reminder by some and as inexplicable by others. Expression design with plants thus calls for careful calibration of legibility and interpretivity \cite{dragan2013legibility, lichtenthaler2016legibility}.

\textbf{Cross-Pollinating Metaphors.} Blending plant gestures with other sources of inspiration can open new creative possibilities. In exploring the expressive potential of our prototypes during our design practice, we found that plant-like gestures often took on unexpected qualities when viewed from an anthropomorphic lens. The \textit{Snake Plant}’s pulsing leaves, for instance, produced a celebratory ``\textit{clapping}'' gesture that participants found delightful. Such hybrid expressions suggest that plant metaphors are not weakened but rather enriched when combined with anthropomorphic echoes.

\textbf{Ambient yet Apparent.} Plant-like expressions walk a delicate line between subtlety and noticeability. Part of their strength often lies in blending into the background, yet gestures that are too muted risk disappearing altogether. Placement amplifies this tension: just as a plant on a dining table becomes a focal presence while one on a shelf recedes into the background, where a robot sits profoundly shapes how its behaviors are seen and read. The \textit{Keeper Plant}’s slow unfurling, for example, was easy to miss unless observed up close, raising concerns in safety-critical contexts like medication management. By contrast, more flamboyant gestures, such as the \textit{Snake Plant}’s fluttering welcome or the \textit{Dancing Pot}’s playful swaying, were warmly received. As P11 reflected, \pquotes{``[when] reading a book or on the computer, you really don’t see much besides what you’re doing... peripheral vision is often lost as you get older.''} emphasizing  the need for careful calibration of scale, timing, and placement to ensure expressions remain ambient without fading into invisibility.

\textbf{Liveliness.} Plant behaviors raise the question of sustaining expressivity over time. The quiet grace of blooming is expressive; however, participants pointed out they could quickly feel repetitive and boring. To stay engaging, robots could echo the variability of natural growth—slight irregularities, subtle shifts, gestures evolving across days or weeks. Approaches such as perlin noise \cite{perlin2002improving, schoen2023lively} or generative models \cite{antony2025xpress} can help here to craft expressions that feel lively in their unfolding, rather than robotically pre-scripted.


\subsection{Rhythms of Expression}
Plants unfold across multiple timescales: rapid nastic movements, daily cycles of opening and closing, and slow seasonal arcs of growth and decay. Such temporal rhythms invite robotic behaviors to range from immediate reactions to long-term ambient signaling. 

\textbf{Art of Timing.} Leveraging plant-like temporality can help mediate responsiveness and intrusiveness in HRI: gradual cues accumulating over minutes or hours may encourage reflection without demanding attention, while more abrupt gestures can be reserved for urgent situations.  P1 illustrated this in updating the Snake Plant imaginary: \pquotes{``slouching for 5 seconds is ok, slouching for 1 hour is a problem. [the leaf] could look weaker and weaker, like its dying, change color [over the hour]... [when no longer slouching] it revives''}. The question is not only when to express, but equally how quickly or slowly and when not, as timing and pacing are critical \cite{nahum2016just}.


\textbf{Layering Gestures.} Experimenting with gestural timescales can enrich expressivity. Depending on how gestures are staged, they can carry different expressive weights: a gentle flutter may register only peripherally, while a dramatic unfurling commands attention. Layering expressive primitives across form and timescale (\eg pairing slow unfolding to reflect long-term progress with brief pulses to mark individual events) can create interactions that are ambient yet meaningful.

\textbf{Non-Reactivity.} The absence of fixed cultural expectations of plant-inspired cues enables them to project presence through subjective interpretations. Their behaviors are open-ended enough to feel supportive rather than evaluative, creating room for quiet companionship rather than constant feedback. As P3 noted \pquotes{``[If I was keeping a mental health voice dairy] it'd be easier to talk to a [plant-like robot] because it wouldn't react, frown on me... how a robot with a face would.''} illustrating how plant-like non-reactivity itself can shape an intentional negative space for gestures. Social context further informs when non-reactivity is desirable: P2 imagined ignoring the Dancing Pot in a group setting and P7 questioned the value of the \textit{Flower Plant} in joining dialogue at all. Robots should know not only what to react to, but also what to let pass in stillness.




\subsection{Shapes of Presence}
Robot morphology is much more than a stylistic choice; it is where context, behavior, and expression converge. Plants provide a guiding metaphor here: their diversity of forms, textures, and scales illustrates the breadth of possibilities for shaping robotic presence.

\textbf{Degrees of Abstraction.} Plant inspiration can be rendered literally, abstractly, or as hybrids. Literal morphologies offer intuitive reading (\eg wilting as distress, blooming as vitality) while abstract or sculptural forms introduce ambiguity that invites interpretation. Participants imagined a plant-like robot doubling as an ambient lamp, or a plant-like interface embedded in digital workflows. In our design practice, we experimented with a leaf module printed in the rectangular form of a ``\textit{robotic bookmark}'' that could unfurl to mark progress in a book and nudge regular reading habit (see Fig. \ref{fig:teaser}). These cases show how expressive potential can be mapped onto diverse forms without requiring literal mimicry.

\textbf{Materiality} Just as textures and colors carry affective weight in plants, material choices shape both lifelikeness and expressive range. In our design practice, we found soft or translucent materials not only foregrounded organic tactility but also enabled playful dynamics (\eg fluttering, curling, pulsing) while rigid or polished finishes reinforced mechanical precision. Participants emphasized that material and aesthetic quality (\eg color choice, finishing) is critical, especially in plant-mimicking robots remarking that artificial plants already achieve convincing realism, and anything less risks breaking believability. Materiality and aesthetics thus become key levers for shaping a robot's presence and acceptance.

\textbf{Scale.} Plants range from small potted companions to towering trees, and their impact shifts with size. Similarly, scale in robotic artifacts conditions how gestures are noticed, interpreted, and valued. Small-scale designs invite intimate engagement: a desktop plant may lean or pulse in ways legible only up close, shaping personal routines. Larger-scale designs command visibility and can serve as focal points in social or public settings, where expressions need to carry across distance. Scale also mediates utility: a lamp-like robotic plant illustrates how larger forms can integrate hybrid functions while still carrying expressive weight, whereas smaller robots may act more as symbolic tokens or ritual partners.

\textbf{Engineering Realities.} While our design practice sketched a broad horizon of possibility, engineering realities delimit how gracefully robots can be manifested. In our practice, we found that actuation noise, torque, motion granularity, heat management, and power requirements all constrained form, presence, and gestures. Prototyping revealed challenges in manufacturing with different materials and ensuring aesthetic quality. These constraints actively shape morphology and thereby presence and behavior, requiring a careful balance between technical feasibility and design intent.
\section{Towards Plant-Inspired Ambient HRI}

Our research-through-design practice explored how plant metaphors can inform HRI by shaping presence, temporality, form, and expression and generated imaginaries for robots that are ambient, gradual, and interpretive. Our design work surfaces important questions about sustainability and responsibility: unlike living plants, robotic artifacts carry large ecological footprint. If successful, plant-like robots may compete with or displace real plants, raising concerns about the broader environmental imaginaries they promote.

We see these tensions not as shortcomings but as critical directions for future work. Further research engaging diverse stakeholders (\eg children, caregivers, behavior-change experts) can  help examine how plant-inspired robots can be situated meaningfully within everyday practices. By embracing these questions alongside plant-inspired design metaphors, future robots may integrate into everyday routines with the quiet grace of botanical life.



\begin{acks}
\textbf{Funding Source:} Johns Hopkins Malone Center for Engineering in Healthcare and National Science Foundation award \#2143704.
\textbf{Author CRediT}: Concept. \& Method. (VNA, CH); Vis. (VNA, ZG, CH); Investigation; Review \& Editing (all); Original Draft (VNA); Funding \& Supervision (CH).
\textbf{AI Statement}: Text edited with LLM; output checked for correctness by authors
\end{acks}

\newpage
\balance
\bibliographystyle{ACM-Reference-Format}
\bibliography{references}

@article{isbister2022design,
  title={Design (not) lost in translation: A case study of an intimate-space socially assistive “robot” for emotion regulation},
  author={Isbister, Katherine and Cottrell, Peter and Cecchet, Alessia and Dagan, Ella and Theofanopoulou, Nikki and Bertran, Ferran Altarriba and Horowitz, Aaron J and Mead, Nick and Schwartz, Joel B and Slovak, Petr},
  journal={ACM Transactions on Computer-Human Interaction (TOCHI)},
  volume={29},
  number={4},
  pages={1--36},
  year={2022},
  publisher={ACM New York, NY}
}

@article{luria2019championing,
  title={Championing Research through design in HRI},
  author={Luria, Michal and Zimmerman, John and Forlizzi, Jodi},
  journal={arXiv preprint arXiv:1908.07572},
  year={2019}
}

@inproceedings{luria2021research,
  title={Research through design approaches in human-robot interaction},
  author={Luria, Michal and Hoggenm{\"u}ller, Marius and Lee, Wen-Ying and Hespanhol, Luke and Jung, Malte and Forlizzi, Jodi},
  booktitle={Companion of the 2021 ACM/IEEE international conference on human-robot interaction},
  pages={685--687},
  year={2021}
}

@inproceedings{hoggenmuller2021eliciting,
  title={Eliciting new perspectives in RtD studies through annotated portfolios: A case study of robotic artefacts},
  author={Hoggenm{\"u}ller, Marius and Lee, Wen-Ying and Hespanhol, Luke and Jung, Malte and Tomitsch, Martin},
  booktitle={Proceedings of the 2021 ACM Designing Interactive Systems Conference},
  pages={1875--1886},
  year={2021}
}

@inproceedings{zimmerman2007research,
  title={Research through design as a method for interaction design research in HCI},
  author={Zimmerman, John and Forlizzi, Jodi and Evenson, Shelley},
  booktitle={Proceedings of the SIGCHI conference on Human factors in computing systems},
  pages={493--502},
  year={2007}
}

@inproceedings{zimmerman2010analysis,
  title={An analysis and critique of Research through Design: towards a formalization of a research approach},
  author={Zimmerman, John and Stolterman, Erik and Forlizzi, Jodi},
  booktitle={proceedings of the 8th ACM conference on designing interactive systems},
  pages={310--319},
  year={2010}
}

@incollection{ljungblad2024critical,
  title={Critical Perspectives in Human--Robot Interaction Design},
  author={Ljungblad, Sara and Gamboa, Mafalda},
  booktitle={Designing Interactions with Robots},
  pages={148--160},
  year={2024},
  publisher={Chapman and Hall/CRC}
}

@inproceedings{winkle2025robots,
  title={Robots from Nowhere: A Case Study in Speculative Sociotechnical Design and Design Fiction for Human-Robot Interaction},
  author={Winkle, Katie},
  booktitle={2025 20th ACM/IEEE International Conference on Human-Robot Interaction (HRI)},
  pages={1152--1165},
  year={2025},
  organization={IEEE}
}

@incollection{zaga2024designing,
  title={Designing Robotic Imaginaries: Narratives and Futures},
  author={Zaga, Cristina},
  booktitle={Designing Interactions with Robots},
  pages={100--131},
  year={2024},
  publisher={Chapman and Hall/CRC}
}

@inproceedings{koike2023exploring,
  title={Exploring the Design Space of Extra-Linguistic Expression for Robots},
  author={Koike, Amy and Mutlu, Bilge},
  booktitle={Proceedings of the 2023 ACM Designing Interactive Systems Conference},
  pages={2689--2706},
  year={2023}
}

@inproceedings{koike2024sprout,
  title={Sprout: Designing expressivity for robots using fiber-embedded actuator},
  author={Koike, Amy and Wehner, Michael and Mutlu, Bilge},
  booktitle={Proceedings of the 2024 ACM/IEEE International Conference on Human-Robot Interaction},
  pages={403--412},
  year={2024}
}

@article{hu2025elegnt,
  title={ELEGNT: Expressive and Functional Movement Design for Non-anthropomorphic Robot},
  author={Hu, Yuhan and Huang, Peide and Sivapurapu, Mouli and Zhang, Jian},
  journal={arXiv preprint arXiv:2501.12493},
  year={2025}
}

@inproceedings{sauppe2014design,
  title={Design patterns for exploring and prototyping human-robot interactions},
  author={Saupp{\'e}, Allison and Mutlu, Bilge},
  booktitle={Proceedings of the SIGCHI conference on human factors in computing systems},
  pages={1439--1448},
  year={2014}
}

@inproceedings{young2007robot,
  title={Robot expressionism through cartooning},
  author={Young, James E and Xin, Min and Sharlin, Ehud},
  booktitle={Proceedings of the ACM/IEEE international conference on Human-robot interaction},
  pages={309--316},
  year={2007}
}

@article{hu2023can,
  title={What can a robot’s skin be? Designing texture-changing skin for human--robot social interaction},
  author={Hu, Yuhan and Hoffman, Guy},
  journal={ACM Transactions on Human-Robot Interaction},
  volume={12},
  number={2},
  pages={1--19},
  year={2023},
  publisher={ACM New York, NY}
}

@inproceedings{singh2012animal,
  title={Animal-inspired human-robot interaction: A robotic tail for communicating state},
  author={Singh, Ashish and Young, James E},
  booktitle={2012 7th ACM/IEEE International Conference on Human-Robot Interaction (HRI)},
  pages={237--238},
  year={2012},
  organization={IEEE}
}

@article{cao2025interruption,
  title={Interruption handling for conversational robots},
  author={Cao, Shiye and Moon, Jiwon and Mahmood, Amama and Antony, Victor Nikhil and Xiao, Ziang and Liu, Anqi and Huang, Chien-Ming},
  journal={arXiv preprint arXiv:2501.01568},
  year={2025}
}

@incollection{ghorbanpour2017importance,
  title={Importance of medicinal and aromatic plants in human life},
  author={Ghorbanpour, Mansour and Hadian, Javad and Nikabadi, Shahab and Varma, Ajit},
  booktitle={Medicinal plants and environmental challenges},
  pages={1--23},
  year={2017},
  publisher={Springer}
}

@article{schaal2019plants,
  title={Plants and people: Our shared history and future},
  author={Schaal, Barbara},
  journal={Plants, People, Planet},
  volume={1},
  number={1},
  pages={14--19},
  year={2019},
  publisher={Wiley Online Library}
}

@article{niazi2023people,
  title={People-plant interaction: plant impact on humans and environment},
  author={Niazi, Parwiz and Alimyar, Obaidullah and Azizi, Azizaqa and Monib, Abdul Wahid and Ozturk, Hamidullah},
  journal={Journal of Environmental and Agricultural Studies},
  volume={4},
  number={2},
  pages={01--07},
  year={2023}
}

@article{antony2025social,
  title={Social Robots for Sleep Health: A Scoping Review},
  author={Antony, Victor and Li, Mengchi and Lin, Shu-Han and Li, Junxin and Huang, Chien-Ming},
  journal={International Journal of Social Robotics},
  pages={1--15},
  year={2025},
  publisher={Springer}
}

@inproceedings{alves2015social,
  title={Social robots for older adults: Framework of activities for aging in place with robots},
  author={Alves-Oliveira, Patr{\'\i}cia and Petisca, Sofia and Correia, Filipa and Maia, Nuno and Paiva, Ana},
  booktitle={International Conference on Social Robotics},
  pages={11--20},
  year={2015},
  organization={Springer}
}

@article{lee2018reframing,
  title={Reframing assistive robots to promote successful aging},
  author={Lee, Hee Rin and Riek, Laurel D},
  journal={ACM Transactions on Human-Robot Interaction (THRI)},
  volume={7},
  number={1},
  pages={1--23},
  year={2018},
  publisher={ACM New York, NY, USA}
}

@inproceedings{sullivan2024making,
  title={Making Informed Decisions: Supporting Cobot Integration Considering Business and Worker Preferences},
  author={Sullivan, Dakota and White, Nathan Thomas and Schoen, Andrew and Mutlu, Bilge},
  booktitle={Proceedings of the 2024 ACM/IEEE International Conference on Human-Robot Interaction},
  pages={706--714},
  year={2024}
}

@inproceedings{kalegina2018characterizing,
  title={Characterizing the design space of rendered robot faces},
  author={Kalegina, Alisa and Schroeder, Grace and Allchin, Aidan and Berlin, Keara and Cakmak, Maya},
  booktitle={Proceedings of the 2018 ACM/IEEE International Conference on Human-Robot Interaction},
  pages={96--104},
  year={2018}
}

@inproceedings{kahn2008design,
  title={Design patterns for sociality in human-robot interaction},
  author={Kahn, Peter H and Freier, Nathan G and Kanda, Takayuki and Ishiguro, Hiroshi and Ruckert, Jolina H and Severson, Rachel L and Kane, Shaun K},
  booktitle={Proceedings of the 3rd ACM/IEEE international conference on Human robot interaction},
  pages={97--104},
  year={2008}
}

@article{sauer2021zoomorphic,
  title={Zoomorphic gestures for communicating cobot states},
  author={Sauer, Vanessa and Sauer, Axel and Mertens, Alexander},
  journal={IEEE Robotics and Automation Letters},
  volume={6},
  number={2},
  pages={2179--2185},
  year={2021},
  publisher={IEEE}
}

@article{kiger2020thematic,
  title={Thematic analysis of qualitative data: AMEE Guide No. 131},
  author={Kiger, Michelle E and Varpio, Lara},
  journal={Medical teacher},
  volume={42},
  number={8},
  pages={846--854},
  year={2020},
  publisher={Taylor \& Francis}
}

@article{antony2025id,
  title={ID. 8: Co-Creating visual stories with Generative AI},
  author={Antony, Victor Nikhil and Huang, Chien-Ming},
  journal={ACM Transactions on Interactive Intelligent Systems},
  volume={14},
  number={3},
  pages={1--29},
  year={2025},
  publisher={ACM New York, NY}
}

@inproceedings{antony2025xpress,
  title={Xpress: A system for dynamic, context-aware robot facial expressions using language models},
  author={Antony, Victor Nikhil and Stiber, Maia and Huang, Chien-Ming},
  booktitle={2025 20th ACM/IEEE International Conference on Human-Robot Interaction (HRI)},
  pages={958--967},
  year={2025},
  organization={IEEE}
}

@ARTICLE{huang_gaze_2015,
author={Huang, Chien-Ming  and Andrist, Sean  and Sauppé, Allison  and Mutlu, Bilge },       
title={Using gaze patterns to predict task intent in collaboration},      
journal={Frontiers in Psychology},       
volume={Volume 6 - 2015},
year={2015},
url={https://www.frontiersin.org/journals/psychology/articles/10.3389/fpsyg.2015.01049},
doi={10.3389/fpsyg.2015.01049},
issn={1664-1078}
}

@article{pfeifer2007self,
  title={Self-organization, embodiment, and biologically inspired robotics},
  author={Pfeifer, Rolf and Lungarella, Max and Iida, Fumiya},
  journal={science},
  volume={318},
  number={5853},
  pages={1088--1093},
  year={2007},
  publisher={American Association for the Advancement of Science}
}

@article{del2024growing,
  title={A growing soft robot with climbing plant--inspired adaptive behaviors for navigation in unstructured environments},
  author={Del Dottore, Emanuela and Mondini, Alessio and Rowe, Nick and Mazzolai, Barbara},
  journal={Science Robotics},
  volume={9},
  number={86},
  pages={eadi5908},
  year={2024},
  publisher={American Association for the Advancement of Science}
}

@inproceedings{yan2019design,
  title={Design of a growing robot inspired by plant growth},
  author={Yan, Tongxi and Teshigawara, Seiichi and Asada, H Harry},
  booktitle={2019 IEEE/RSJ International Conference on Intelligent Robots and Systems (IROS)},
  pages={8006--8011},
  year={2019},
  organization={IEEE}
}

@article{speck2023plants,
  title={Plants as inspiration for material-based sensing and actuation in soft robots and machines},
  author={Speck, Thomas and Cheng, Tiffany and Klimm, Frederike and Menges, Achim and Poppinga, Simon and Speck, Olga and Tahouni, Yasaman and Tauber, Falk and Thielen, Marc},
  journal={MRS Bulletin},
  volume={48},
  number={7},
  pages={730--745},
  year={2023},
  publisher={Springer}
}

@article{jun2017plant,
  title={Plant-like robots.},
  author={Jun, Ji Won},
  journal={Interactions},
  volume={24},
  number={5},
  pages={88},
  year={2017}
}

@inproceedings{hu2024designing,
  title={Designing plant-driven actuators for robots to grow, age, and decay},
  author={Hu, Yuhan and Lu, Jasmine and Scinto-Madonich, Nathan and Pineros, Miguel Alfonso and Lopes, Pedro and Hoffman, Guy},
  booktitle={Proceedings of the 2024 ACM Designing Interactive Systems Conference},
  pages={2481--2496},
  year={2024}
}

@inproceedings{holstius2004infotropism,
  title={Infotropism: living and robotic plants as interactive displays},
  author={Holstius, David and Kembel, John and Hurst, Amy and Wan, Peng-Hui and Forlizzi, Jodi},
  booktitle={Proceedings of the 5th conference on Designing interactive systems: processes, practices, methods, and techniques},
  pages={215--221},
  year={2004}
}

@article{bhat2021plant,
  title={Plant robot for at-home behavioral activation therapy reminders to young adults with depression},
  author={Bhat, Ashwin Sadananda and Boersma, Christiaan and Meijer, Max Jan and Dokter, Maaike and Bohlmeijer, Ernst and Li, Jamy},
  journal={ACM Transactions on Human-Robot Interaction (THRI)},
  volume={10},
  number={3},
  pages={1--21},
  year={2021},
  publisher={ACM New York, NY, USA}
}

@inproceedings{grandos2025roboplant,
author = {Granados, Carlos and De la Cruz, Katherine and Tafur, Mayli and Campos, Miguel and Chavez, Manuel and Arce, Diego},
title = {PlantiBot: Towards the Design of a Robotic Plant for Mental Health Care},
year = {2025},
isbn = {9798400711978},
publisher = {Association for Computing Machinery},
address = {New York, NY, USA},
url = {https://doi.org/10.1145/3689050.3705996},
doi = {10.1145/3689050.3705996},
booktitle = {Proceedings of the Nineteenth International Conference on Tangible, Embedded, and Embodied Interaction},
articleno = {81},
numpages = {6},
location = {
},
series = {TEI '25}
}

@article{mo2020jumping,
  title={Jumping locomotion strategies: From animals to bioinspired robots},
  author={Mo, Xiaojuan and Ge, Wenjie and Miraglia, Marco and Inglese, Francesco and Zhao, Donglai and Stefanini, Cesare and Romano, Donato},
  journal={Applied Sciences},
  volume={10},
  number={23},
  pages={8607},
  year={2020},
  publisher={MDPI}
}

@article{rahmani2016robust,
  title={Robust adaptive control of a bio-inspired robot manipulator using bat algorithm},
  author={Rahmani, Mehran and Ghanbari, Ahmad and Ettefagh, Mir Mohammad},
  journal={Expert Systems with Applications},
  volume={56},
  pages={164--176},
  year={2016},
  publisher={Elsevier}
}

@inproceedings{nertinger2024designing,
  title={Designing Humanoids: How Robot Posture Influences Users’ Perceived Safety in HRI},
  author={Nertinger, Simone and Herzog, Olivia and M{\"u}hlbauer, Anna and Naceri, Abdeldjallil and Haddadin, Sami},
  booktitle={2024 IEEE-RAS 23rd International Conference on Humanoid Robots (Humanoids)},
  pages={45--52},
  year={2024},
  organization={IEEE}
}

@article{gao2022evaluation,
  title={Evaluation of socially-aware robot navigation},
  author={Gao, Yuxiang and Huang, Chien-Ming},
  journal={Frontiers in Robotics and AI},
  volume={8},
  pages={721317},
  year={2022},
  publisher={Frontiers Media SA}
}

@inproceedings{bruce2002role,
  title={The role of expressiveness and attention in human-robot interaction},
  author={Bruce, Allison and Nourbakhsh, Illah and Simmons, Reid},
  booktitle={Proceedings 2002 IEEE international conference on robotics and automation (Cat. No. 02CH37292)},
  volume={4},
  pages={4138--4142},
  year={2002},
  organization={IEEE}
}

@inproceedings{Lindley2020AIDesign,
author = {Lindley, Joseph and Akmal, Haider Ali and Pilling, Franziska and Coulton, Paul},
title = {Researching AI Legibility through Design},
year = {2020},
isbn = {9781450367080},
publisher = {Association for Computing Machinery},
address = {New York, NY, USA},
url = {https://doi.org/10.1145/3313831.3376792},
doi = {10.1145/3313831.3376792},
booktitle = {Proceedings of the 2020 CHI Conference on Human Factors in Computing Systems},
pages = {1–13},
numpages = {13},
location = {Honolulu, HI, USA},
series = {CHI '20}
}

@inproceedings{Benjamin2024GenAIRtD,
author = {Benjamin, Jesse Josua and Lindley, Joseph and Edwards, Elizabeth and Rubegni, Elisa and Korjakow, Tim and Grist, David and Sharkey, Rhiannon},
title = {Responding to Generative AI Technologies with Research-through-Design: The Ryelands AI Lab as an Exploratory Study},
year = {2024},
isbn = {9798400705830},
publisher = {Association for Computing Machinery},
address = {New York, NY, USA},
url = {https://doi.org/10.1145/3643834.3660677},
doi = {10.1145/3643834.3660677},
booktitle = {Proceedings of the 2024 ACM Designing Interactive Systems Conference},
pages = {1823–1841},
numpages = {19},
location = {Copenhagen, Denmark},
series = {DIS '24}
}

@inproceedings{sengers2005reflective,
  title={Reflective design},
  author={Sengers, Phoebe and Boehner, Kirsten and David, Shay and Kaye, Joseph'Jofish'},
  booktitle={Proceedings of the 4th decennial conference on Critical computing: between sense and sensibility},
  pages={49--58},
  year={2005}
}

@article{lowgren2013annotated,
  title={Annotated portfolios and other forms of intermediate-level knowledge},
  author={L{\"o}wgren, Jonas},
  journal={interactions},
  volume={20},
  number={1},
  pages={30--34},
  year={2013},
  publisher={ACM New York, NY, USA}
}

@inproceedings{lupetti2021designerly,
  title={Designerly ways of knowing in HRI: Broadening the scope of design-oriented HRI through the concept of intermediate-level knowledge},
  author={Lupetti, Maria Luce and Zaga, Cristina and Cila, Nazli},
  booktitle={Proceedings of the 2021 ACM/IEEE International Conference on Human-Robot Interaction},
  pages={389--398},
  year={2021}
}

@article{segura2016bodystorming,
  title={Bodystorming for movement-based interaction design},
  author={Segura, Elena and Vidal, Laia and Rostami, Asreen},
  journal={Human Technology},
  volume={12},
  number={2},
  pages={193--251},
  year={2016}
}

@inproceedings{antony2025onbodydesign,
  title={The design of on-body robots for older adults},
  author={Antony, Victor Nikhil and Jeon, Clara and Li, Jiasheng and Gao, Ge and Peng, Huaishu and Ostrowski, Anastasia K and Huang, Chien-Ming},
  booktitle={2025 20th ACM/IEEE International Conference on Human-Robot Interaction (HRI)},
  pages={589--598},
  year={2025},
  organization={IEEE}
}

@inproceedings{tang2022confidant,
  title={Confidant: A privacy controller for social robots},
  author={Tang, Brian and Sullivan, Dakota and Cagiltay, Bengisu and Chandrasekaran, Varun and Fawaz, Kassem and Mutlu, Bilge},
  booktitle={2022 17th ACM/IEEE International Conference on Human-Robot Interaction (HRI)},
  pages={205--214},
  year={2022},
  organization={IEEE}
}

@article{rueben2016privacy,
  title={Privacy in human-robot interaction: Survey and future work},
  author={Rueben, Matthew and Smart, William D},
  journal={We robot},
  volume={2016},
  pages={5th},
  year={2016}
}

@inproceedings{karli2024alchemist,
  title={Alchemist: Llm-aided end-user development of robot applications},
  author={Karli, Ulas Berk and Chen, Juo-Tung and Antony, Victor Nikhil and Huang, Chien-Ming},
  booktitle={Proceedings of the 2024 ACM/IEEE International Conference on Human-Robot Interaction},
  pages={361--370},
  year={2024}
}

@article{glas2012network,
  title={The network robot system: enabling social human-robot interaction in public spaces},
  author={Glas, Dylan F and Satake, Satoru and Ferreri, Florent and Kanda, Takayuki and Hagita, Norihiro and Ishiguro, Hiroshi},
  journal={Journal of Human-Robot Interaction},
  volume={1},
  number={2},
  pages={5--32},
  year={2012}
}

@inproceedings{zhao2018through,
  title={Through-wall human pose estimation using radio signals},
  author={Zhao, Mingmin and Li, Tianhong and Abu Alsheikh, Mohammad and Tian, Yonglong and Zhao, Hang and Torralba, Antonio and Katabi, Dina},
  booktitle={Proceedings of the IEEE conference on computer vision and pattern recognition},
  pages={7356--7365},
  year={2018}
}

@article{nahum2016just,
  title={Just-in-time adaptive interventions (JITAIs) in mobile health: key components and design principles for ongoing health behavior support},
  author={Nahum-Shani, Inbal and Smith, Shawna N and Spring, Bonnie J and Collins, Linda M and Witkiewitz, Katie and Tewari, Ambuj and Murphy, Susan A},
  journal={Annals of behavioral medicine},
  pages={1--17},
  year={2016},
  publisher={Springer}
}

@inproceedings{dragan2013legibility,
  title={Legibility and predictability of robot motion},
  author={Dragan, Anca D and Lee, Kenton CT and Srinivasa, Siddhartha S},
  booktitle={2013 8th ACM/IEEE International Conference on Human-Robot Interaction (HRI)},
  pages={301--308},
  year={2013},
  organization={IEEE}
}

@article{lichtenthaler2016legibility,
  title={Legibility of robot behavior: A literature review},
  author={Lichtenth{\"a}ler, Christina and Kirsch, Alexandra},
  year={2016}
}

@inproceedings{perlin2002improving,
  title={Improving noise},
  author={Perlin, Ken},
  booktitle={Proceedings of the 29th annual conference on Computer graphics and interactive techniques},
  pages={681--682},
  year={2002}
}

@inproceedings{schoen2023lively,
  title={Lively: Enabling multimodal, lifelike, and extensible real-time robot motion},
  author={Schoen, Andrew and Sullivan, Dakota and Zhang, Ze Dong and Rakita, Daniel and Mutlu, Bilge},
  booktitle={Proceedings of the 2023 ACM/IEEE International Conference on Human-Robot Interaction},
  pages={594--602},
  year={2023}
}

@inproceedings{antony2023co,
  title={Co-designing with older adults, for older adults: robots to promote physical activity},
  author={Antony, Victor Nikhil and Cho, Sue Min and Huang, Chien-Ming},
  booktitle={Proceedings of the 2023 ACM/IEEE International Conference on Human-Robot Interaction},
  pages={506--515},
  year={2023}
}

@inproceedings{mott2022robot,
  title={Robot co-design can help us engage child stakeholders in ethical reflection},
  author={Mott, Terran and Bejarano, Alexandra and Williams, Tom},
  booktitle={2022 17th ACM/IEEE International Conference on Human-Robot Interaction (HRI)},
  pages={14--23},
  year={2022},
  organization={IEEE}
}

@inproceedings{alves2021collection,
  title={Collection of metaphors for human-robot interaction},
  author={Alves-Oliveira, Patr{\'\i}cia and Lupetti, Maria Luce and Luria, Michal and L{\"o}ffler, Diana and Gamboa, Mafalda and Albaugh, Lea and Kamino, Waki and K. Ostrowski, Anastasia and Puljiz, David and Reynolds-Cu{\'e}llar, Pedro and others},
  booktitle={Proceedings of the 2021 ACM Designing Interactive Systems Conference},
  pages={1366--1379},
  year={2021}
}

@inproceedings{ostrowski2020design,
  title={Design research in HRI: roboticists, design features, and users as co-designers},
  author={Ostrowski, Anastasia K and Park, Hae Won and Breazeal, Cynthia},
  booktitle={Workshop on Designerly HRI Knowledge},
  year={2020}
}

@inproceedings{hook2016somaesthetic,
  title={Somaesthetic appreciation design},
  author={H{\"o}{\"o}k, Kristina and Jonsson, Martin P and St{\aa}hl, Anna and Mercurio, Johanna},
  booktitle={Proceedings of the 2016 chi conference on human factors in computing systems},
  pages={3131--3142},
  year={2016}
}

@inproceedings{thompson2021ambidots,
  title={Ambidots: An ambient interface to mediate casual social settings through peripheral interaction},
  author={Thompson, Edward and Potts, Dominic and Hardy, John and Porter, Barry and Houben, Steven},
  booktitle={Proceedings of the 33rd Australian Conference on Human-Computer Interaction},
  pages={99--110},
  year={2021}
}

@article{siniscalchi2018communication,
  title={Communication in dogs},
  author={Siniscalchi, Marcello and d’Ingeo, Serenella and Minunno, Michele and Quaranta, Angelo},
  journal={Animals},
  volume={8},
  number={8},
  pages={131},
  year={2018},
  publisher={MDPI}
}

\appendix

\end{document}